\documentclass{article}

     \usepackage[preprint, nonatbib]{neurips_2020}
    \usepackage{appendix}
\usepackage{todonotes}
\usepackage[utf8]{inputenc} 
\usepackage[T1]{fontenc}    
\usepackage{hyperref}       
\usepackage{url}            
\usepackage{booktabs}       
\usepackage{amsfonts}       
\usepackage{nicefrac}       
\usepackage{microtype}      
\usepackage{amsmath}
\usepackage{caption}
\usepackage{subcaption}
\usepackage[ruled,vlined]{algorithm2e}
\usepackage{enumitem}
\usepackage{bm}

\title{Robust Reinforcement Learning \\ using Adversarial Populations}

\author{%
  Eugene Vinitsky\thanks{Equal authorship} \\
  Department of Mechanical Engineering\\
  UC Berkeley\\
  \texttt{evinitsky@berkeley.edu} \\
  \And
  Yuqing Du\footnotemark[1] \\
  Department of EECS\\
 UC Berkeley\\
  \texttt{yuqing\_du@berkeley.edu} \\
    \And
  Kanaad Parvate\footnotemark[1]  \\
  Department of EECS\\
 UC Berkeley\\
  \texttt{kanaad@berkeley.edu} \\
    \And
  Kathy Jang \\
  Department of EECS\\
 UC Berkeley\\
  \texttt{kathyjang@berkeley.edu} \\
    \And
  Pieter Abbeel \\
  Department of EECS\\
 UC Berkeley\\
  \texttt{pabbeel@cs.berkeley.edu} \\
    \And
  Alexandre Bayen \\
  Department of EECS\\
  Institute of Transportation Studies\\
 UC Berkeley\\
  \texttt{bayen@berkeley.edu} \\
}

\begin{document}

\maketitle

\begin{abstract}
 Reinforcement Learning (RL) is an effective tool for controller design but can struggle with issues of robustness, failing catastrophically when the underlying system dynamics are perturbed. The Robust RL formulation tackles this by adding worst-case adversarial noise to the dynamics and constructing the noise distribution as the solution to a zero-sum minimax game. However, existing work on learning solutions to the Robust RL formulation has primarily focused on training a single RL agent against a single adversary. In this work, we demonstrate that using a single adversary does not consistently yield robustness to dynamics variations under standard parametrizations of the adversary; the resulting policy is highly exploitable by new adversaries. We propose a population-based augmentation to the Robust RL formulation in which we randomly initialize a population of adversaries and sample from the population uniformly during training. We empirically validate across robotics benchmarks that the use of an adversarial population results in a more robust policy that also improves out-of-distribution generalization. Finally, we demonstrate that this approach provides comparable robustness and generalization as domain randomization on these benchmarks while avoiding a ubiquitous domain randomization failure mode. 
 

\end{abstract}

\section{Introduction}
\label{sec:introduction}
Developing controllers that work effectively across a wide range of potential deployment environments is one of the core challenges in engineering. The complexity of the physical world means that the models used to design controllers are often inaccurate. Optimization based control design approaches, such as reinforcement learning (RL), have no notion of model inaccuracy and can lead to controllers that fail catastrophically under mismatch. In this work, we aim to demonstrate an effective method for training reinforcement learning policies that are robust to model inaccuracy by designing controllers that are effective in the presence of worst-case adversarial noise in the dynamics.

One effective approach to induce robustness has been domain randomization \cite{tobin2017domain, jakobi1997evolutionary}, a method where a designer with expertise identifies the components of the model that they are uncertain about. They then construct a set of training environments where the uncertain components are randomized, ensuring that the agent is robust on average to this set. However, this requires careful parametrization of the uncertainty set as well as hand-designing of the environments.

A more easily automated approach is to formulate the problem as a zero-sum game and learn an adversary that perturbs the transition dynamics~\cite{tessler2019action, kamalaruban2020robust, pinto2017robust}. If a global Nash equilibrium of this problem is found, then that equilibrium provides a worst case performance bound under the specified set of perturbations. Besides the benefit of removing user design once the perturbation mechanism is specified, this approach is maximally conservative, which is useful for safety critical applications.

However, the aforementioned literature on learning an adversary predominantly uses a single stochastic adversary. This raises a puzzling question: the minimax problem does not necessarily have any pure Nash equilibria (see Appendix C~\cite{tessler2019action}) but the existing robust RL literature mostly appears to attempt to solve for pure Nash equilibria. That is, the most general form of the minimax problem searches over distributions of adversary and agent policies
\begin{equation}
\label{eq:full_adv_dynamics}
\underset{p \in \mathcal{P}(\Theta)}{\text{max}} \underset{q \in \mathcal{Q}(\Omega)}{\text{min}} \mathbb{E}_{\theta \sim p} \left[ \mathbb{E}_{\omega \sim q} \left[h(\theta, \omega) \right]\right]
\end{equation}
where $\mathcal{P}(\Theta), \mathcal{Q}(\Omega)$ are distributions over policies and $h(\theta, \omega)$ is a score function (for example, expected cumulative reward). However, this problem is approximated in the literature by the fixed-policy problem 
\begin{equation}
\label{eq:simple_adv_dynamics}
\underset{\theta}{\text{max}} \ \  \underset{\omega}{\text{min}}\ \  h(\theta, \omega)
\end{equation}
We contend that this reduction to a single adversary approach can sometimes fail to result in improved robustness under standard parametrizations of the adversary policy.




The following example provides some intuition for why using a single adversary can decrease robustness. Consider a robot trying to learn to walk east-wards while an adversary outputs a force representing wind coming from the north or the south. For a fixed, deterministic adversary the agent knows that the wind will come from either south or north and can simply apply a counteracting force at each state. Once the adversary is removed, the robot will still apply the compensatory forces and possibly become unstable. Stochastic Gaussian policies (which are ubiquitous in continuous control) offer little improvement: low entropy policies can be counteracted whereas high entropy policies would endow the robot with the prior that the wind cancels on average. Under these standard policy parametrizations, which cannot represent a distribution over policies, we cannot use an adversary to endow the agent with a prior that a persistent, strong wind could come either from north or south. This leaves the agent exploitable to this class of perturbations.

The use of a single adversary in the robustness literature is in contrast to the multi-player game literature. In multi-player games, large sets of adversaries are used to ensure that an agent cannot easily be exploited~\cite{vinyals2019grandmaster, czarnecki2020real, brown2019superhuman}. Drawing inspiration from this literature, we introduce \textbf{RAP} (Robustness via Adversary Populations): a randomly initialized population of adversaries that we sample from at each rollout and train alongside the agent. Returning to our example of a robot perturbed by wind, if the robot learns to cancel any one of the adversaries effectively, then that opens a niche for an adversary to exploit by applying forces in another direction. As the number of adversaries increases, the robot is eventually endowed with the prior that a strong wind could come from either direction and that it must walk carefully to avoid being toppled over.


Our contributions are as follows:
\begin{itemize}
    \item Using a set of continuous control tasks, we provide evidence that a single adversary does not have a consistent positive impact on the robustness of an RL policy while the use of an adversary population provides improved robustness across all considered examples. 
    \item We investigate the source of the robustness and show that the single adversary policy is exploitable by new adversaries whereas policies trained with RAP are robust to new adversaries.
    \item We demonstrate that adversary populations can be competitive with domain randomization while avoiding potential failure modes of domain randomization.
\end{itemize}

\section{Related Work}
\vspace{-0.2cm}
This work builds upon robust control \cite{zhou1998essentials}, a branch of control theory focused on finding optimal controllers under worst-case perturbations of the system dynamics. The Robust Markov Decision Process (R-MDP) formulation extends this worst-case model uncertainty to uncertainty sets on the transition dynamics of an MDP and demonstrates that computationally tractable solutions exist for small, tabular MDPs~\cite{nilim2005robust, lim2013reinforcement}. For larger or continuous MDPs, one successful approach has been to use function approximation to compute approximate solutions to the R-MDP problem~\cite{tamar2014scaling}. 

One prominent variant of the R-MDP literature is to interpret the perturbations as an adversary and attempt to learn the distribution of the perturbation under a minimax objective. Two variants of this idea that tie in closely to our work are Robust Adversarial Reinforcement Learning (RARL)~\cite{pinto2017robust}  and and Noisy Robust Markov Decision Processes (NR-MDP)~\cite{tessler2019action} which differ in how they parametrize the adversaries: RARL picks out specific robot joints that the adversary acts on while NR-MDP adds the adversary action to the agent action. Both of these works attempt to find an equilibrium of the minimax objective using a single adversary; in contrast our work uses a large set of adversaries and shows improved robustness relative to a single adversary.

An alternative to the minimax objective, domain randomization, asks a designer to explicitly define a distribution over environments that the agent should be robust to. For example,~\cite{peng2018sim} varies simulator friction, mass, table height, and controller gain (along with several other parameters) to train a robot to robustly push a puck to a target location in the real world; ~\cite{antonova2017reinforcement} added noise to friction and actions to transfer an object pivoting policy directly from simulation to a Baxter robot. Additionally, domain randomization has been successfully used to build accurate object detectors solely from simulated data~\cite{tobin2017domain}, to zero-shot transfer a quadcopter flight policy from simulation ~\cite{sadeghi2016cad2rl}.

However, as we discuss in Sec.~\ref{sec:mujoco_results}, a policy that performs well on average across simulation domains is not necessarily robust as it may trade off performance on one set of parameters to maximize performance in another. EPOpt~\cite{rajeswaran2016epopt} addresses this by replacing the uniform average across distributions with the conditional value at risk (CVaR)~\cite{chow2015risk, tamar2015optimizing} a soft version of the minimax objective in which the optimization is only performed over a small percentage of the worst performing parameters. This is an interesting approach to align the domain randomization objective with the minimax objective and could be made compatible with our approach by only training using a subset of the strongest adversaries. 

Our demonstration of overfitting to a single adversary is not new; there is extensive work establishing that agents trained independently in multi-agent settings can result in non-robust policies.~\cite{gleave2019adversarial} show that in zero-sum games adversary pairs trained via RL are not robust to replacement of the adversary with a different adversary policy. ~\cite{lanctot2017unified} extends this idea to general sum games by training a population of agent-agent pairs and showing that taking two pairs and swapping the agents in them leads to failure to accomplish the objective.~\cite{shapley1964some} establishes that even in tabular settings (in this case, a general-sum version of Rock Paper Scissors), iterated best response to pure Nash strategies can lead to cyclical behavior and a failure to converge to equilibrium. 

The use of population based training is also a standard technique in multi-agent settings. Alphastar, the grandmaster-level Starcraft bot, uses a population of "exploiter" agents that fine-tune against the bot to prevent it from developing exploitable strategies~\cite{vinyals2019grandmaster}.~\cite{czarnecki2020real} establishes a set of sufficient geometric conditions on games under which the use of multiple adversaries will ensure gradual improvement in the strength of the agent policy. They empirically demonstrate that learning in games can often fail to converge without populations. Finally, Active Domain Randomization~\cite{mehta2019active} is a very close approach to ours, as they use a population of adversaries to select domain randomization parameters whereas we use a population of adversaries to directly perturb the agent actions. Additionally, they use a Stein Variation Policy Gradient~\cite{liu2017stein} to ensure diversity in their adversaries and a discriminator reward instead of a minimax reward whereas our work does not have any explicit coupling between the adversary gradient updates and uses a simpler zero-sum reward function.

\section{Background}

\subsection{Notation.}
In this work we use the framework of a multi-agent, finite-horizon, discounted, Markov Decision Process (MDP) \cite{puterman1990markov} defined by a tuple $\langle A_\text{agent} \times A_\text{adversary}, S, \mathcal{T}, \mathcal{R}, \gamma \rangle$. Here $A_\text{agent}$ is the set of actions for the agent, $A_\text{adversary}$ is the set of actions for the adversary, $S$ is a set of states, $\mathcal{T}: A_\text{agent} \times A_\text{adversary} \times S \rightarrow \Delta(S)$ is a transition function, $R: A_\text{agent} \times A_\text{adversary} \times S \rightarrow \mathbb{R}$ is a reward function and $\gamma$ is a discount factor. $S$ is shared between the adversaries as they share a state-space with the agent. The goal for a given MDP is to find a policy $\pi_\theta$ parametrized by $\theta$ that maximizes the expected cumulative discounted reward $J^\theta = \mathbb{E} \left[\sum_{t=0}^T \gamma^t r(s_t, a_t) \vert \pi_\theta \right]$. The conditional in this expression is a short-hand to indicate that the actions in the MDP are sampled via $a_t \sim \pi_\theta(s_t, a_{t-1})$. We denote the agent policy parametrized by weights $\theta$ as $\pi_\theta$ and the policy of adversary $i$ as $\bar{\pi}_{\phi_i}$.  Actions sampled from the adversary policy $\bar{\pi}_{\phi_i}$ will be written as $\bar{a}_t^i$. We use $\xi$ to denote the parametrization of the system dynamics (e.g. different values of friction, mass, wind, etc.) and the system dynamics for a given state and action as $s_{t+1} \sim f_\xi(s_t, a_t)$.

\subsection{Baselines}
Here we outline prior work and the approaches that will be compared with RAP. Our baselines consist of a single adversary and domain randomization.

\subsubsection{Single Minimax Adversary}
\label{sec:minimax_adversary}

Our adversary formulation uses the \emph{Noisy Action Robust MDP} ~\cite{tessler2019action} in which the adversary adds its actions onto the agent actions. The objective is 

\begin{equation}
\label{eq:min_max_definition_action}
\begin{gathered}
\underset{\theta}{\text{max}} \
    \underset{\phi}{\text{min}} \   \mathbb{E} \left[\sum_{t=0}^T \gamma^t r(s_t, a_t + \alpha \bar{a_t}) | \pi_\theta, \ \bar{\pi}_{\phi} \right]\, \\
\end{gathered}
\end{equation}
where $\alpha$ is a hyperparameter controlling the adversary strength.

We note two important restrictions inherent to this adversarial model. First, since the adversary is only able to attack the agent through the actions, there is a restricted class of dynamical systems that it can represent; this set of dynamical systems may not necessarily align with the set of dynamical systems that the agent may be tested in. This is simply a restriction caused by the choice of adversarial perturbation and could be alleviated by using different adversarial parametrizations e.g. perturbing the transition function directly.

In addition to the restricted set of dynamical systems that the NR-MDP can represent, there is a limitation induced by standard RL agent parametrizations. In particular, agents are often parametrized by either having deterministic actions or having their actions drawn from a probability distribution (i.e. we pass a state through our policy, it outputs parameters of a distribution and we sample the actions from that distribution). The single adversary cannot represent all the systems that we intend the agent to be robust to as a consequence of the parametrization. For example, suppose the agent is currently at some state $s_t$ and the adversary outputs an action $\bar{a_t}$. In the deterministic case, the agent knows that the adversary will never output $-\bar{a_t}$ even though $-\bar{a_t}$ is clearly in the class of possible perturbations. 



\subsubsection{Dynamics Randomization}\label{sec:dr}


Domain randomization is the setting in which the user specifies a set of environments which the agent should be robust to. This allows the user to directly encode knowledge about the likely deviations between training and testing domains. For example, the user may believe that friction is hard to measure precisely and wants to ensure that their agent is robust to variations in friction; they then specify that the agent will be trained with a wide range of possible friction values. We use $\xi$ to denote some vector that parametrizes the set of training environments (e.g. friction, masses, system dynamics, etc.). 
We denote the domain over which $\xi$ is drawn from as $\Xi$ and use $\mathcal{P}\left(\Xi\right)$ to denote some probability distribution over $\xi$. The domain randomization objective is
\begin{equation}
\begin{gathered}
\label{eq:domain_randomization}
\underset{\theta}{\text{max}} \ \mathbb{E}_{\xi \sim \mathcal{P}\left(\Xi \right)}\left[\mathbb{E}_{s_{t+1} \sim f_\xi(s_t, a_t)} \left[\sum_{t=0}^T \gamma^t r(s_t, a_t)  | \pi_\theta\right]\right] \\
s_{t+1} \sim f_\xi(s_t, a_t) \\
a_t \sim \pi_\theta (s_t)
\end{gathered}
\end{equation}
Here the goal is to find an agent that performs well on average across the distribution of training environment.
Most commonly, and in this work, the parameters $\xi$ are sampled uniformly over $\Xi$.

\section{RAP: Robustness via Adversary Populations} \textbf{RAP} extends the minimax objective with a population based approach. Instead of a single adversary, at each rollout we will sample uniformly from a population of adversaries. By using a population, the agent is forced to be robust to a wide variety of potential perturbations instead of a single perturbation.
If the agent begins to overfit to any one adversary, this opens up a potential niche for another adversary to exploit. 
For problems with only one failure mode, we expect the adversaries to all come out identical to the minimax adversary, but as the number of failure modes increases the adversaries should begin to diversify to exploit the agent.  To induce this diversity, we will rely on randomness in the gradient estimates and randomness in the initializations of the adversary networks rather than any explicit term that induces diversity. While the idea of using populations does not preclude explicit terms in the loss to encourage diversity, we find that our chosen sources of diversity are sufficient for our purposes.

Denoting $\bar{\pi}_{\phi_i}$ as the $i$-th adversary and $i \sim U(1, n)$ as the discrete uniform distribution defined on 1 through n, the objective becomes
\begin{equation}
\begin{gathered}
\label{eq:min_max_definition_start}
    \underset{\theta}{\text{max}} \,
    \underset{\phi_1}{\text{min}} \dots \underset{\phi_n}{\text{min}}  \ \    \ \mathbb{E}_{i \sim U(1,n)}  \left[\sum_{t=0}^T \gamma^t r(s_t, a_t, \alpha \bar{a}_t^i) | \pi_\theta, \ \bar{\pi_{\phi_i}} \right]
    \\
    s_{t+1} \sim f(s_t, a_t + \alpha \bar{a_t}) \\
\end{gathered}
\end{equation}
For a single adversary, this is equivalent to the \emph{minimax adversary} described in Sec.~\ref{sec:minimax_adversary}

We will optimize this objective by converting the problem into the equivalent zero-sum game. At the start of each rollout, we will sample an adversary index from the uniform distribution and collect a trajectory in using the agent and the selected adversary. For notational simplicity, we assume the trajectory is of length M and that adversary $i$ will participate in $J_i$ total trajectories while, since the agent participates in every rollout, it will receive J total trajectories. We denote the j-th collected trajectory for the agent as $\tau_j = (s_0, a_0, r_0, s_1) \times \dots  \times (s_M, a_M, r_M, s_{M+1})$ and the associated trajectory for adversary $i$ as $\tau^i_j = (s_0, a_0, -r_0, s_1) \times \dots  \times (s_M, a_M, -r_M, s_M)$. Note that the adversary reward is simply the negative of the agent reward. 

We will use Proximal Policy Optimization~\cite{schulman2017proximal} (PPO) to update our policies.
We caution that we have overloaded notation slightly here and for adversary $i$, $\tau^i_{j=1:J_i}$ refers only to the trajectories in which the adversary was selected: adversaries will only be updated using trajectories where they were active. At the end of a training iteration, we update all our policies using gradient descent. The algorithm is summarized below:

\begin{algorithm}[H]
\label{algo:RAP}
\SetAlgoLined
 Initialize $\theta, \phi_1 \cdots \phi_n$ using Xavier initialization \cite{glorot2010understanding}\;
 \While{not converged}{
    \For{rollout j=1...J}{
        sample adversary $i \sim U(1,n)$\;
        run policies $\pi_{\theta}, \bar{\pi_{\phi_i}}$ in environment until termination\; collect trajectories $\tau_j$, $\tau_j^i$
    }
    update $\theta, \phi_1 \cdots \phi_n$ using PPO \cite{schulman2017proximal} and the collected trajectories\;
 }
 \caption{Robustness via Adversary Populations}
\end{algorithm}

We call the agent trained to optimize this objective using Algorithm \ref{algo:RAP} the \emph{RAP agent}.

\section{Experiments}
In this section we present experiments on continuous control tasks from the OpenAI Gym Suite \cite{openaigym, todorov2012mujoco}. We compare with our baselines and evaluate the efficacy of a population of learned adversaries across a wide range of state and action space sizes. We investigate the following hypotheses:
\begin{enumerate}[label=H\arabic*.]
    \item \label{H1} Agents are more likely to overfit to a single adversary than a population of adversaries, leaving them more exploitable on in-distribution tasks. 
    \item \label{H2} Agents trained against a population of adversaries will generalize better, leading to improved performance on out-of-distribution tasks. 
    \item \label{H3}Naive parametrization of domain randomization can result in a brittle policy, even when evaluated on the same distribution it was trained on.
    \item \label{H4}While a larger adversary populations can represent more varied dynamics, there will be diminishing returns due to the decreased environment steps each adversary receives.
\end{enumerate}

In-distribution tasks refer to the agent playing against perturbations that are in the training distribution: adversaries that add their actions onto the agent. However, the particular form of the adversary and their restricted perturbation magnitude means that there are many dynamical systems that they cannot represent (for example, significant variations of joint mass and friction). These tasks are denoted as out of distribution tasks. All of the tasks in the test set described in Sec.~\ref{sec:exp_setup} are likely out-of-distribution tasks.

\subsection{Experimental Setup and Hyperparameter Selection}\label{sec:exp_setup}
While we provide exact details of the hyperparameters in the Appendix, adversarial settings require additional complexity in hyperparameter selection. In the standard RL procedure, optimal hyperparameters are selected on the basis of maximum expected cumulative reward. However, if an agent playing against an adversary achieves a large cumulative reward, it is possible that the agent was simply playing against a weak adversary. Conversely, a low score does not necessarily indicate a strong adversary nor robustness: it could simply mean that we trained a weak agent. 

To address this, we adopt a version of the train-validate-test split from supervised learning. We use the mean policy performance on a suite of validation tasks to select the hyperparameters, then we train the policy across ten seeds and report the resultant mean and standard deviation over twenty trajectories. Finally, we evaluate the seeds on a holdout test set of eight additional model-mismatch tasks. These tasks vary significantly in difficulty; for visual clarity we report only the average across tasks in this paper and report the full breakdown across tasks in the Appendix.

\begin{figure}
\begin{subfigure}[b]{\textwidth}
     \centering
     \begin{subfigure}[b]{0.3\textwidth}
         \centering
         \includegraphics[width=\textwidth]{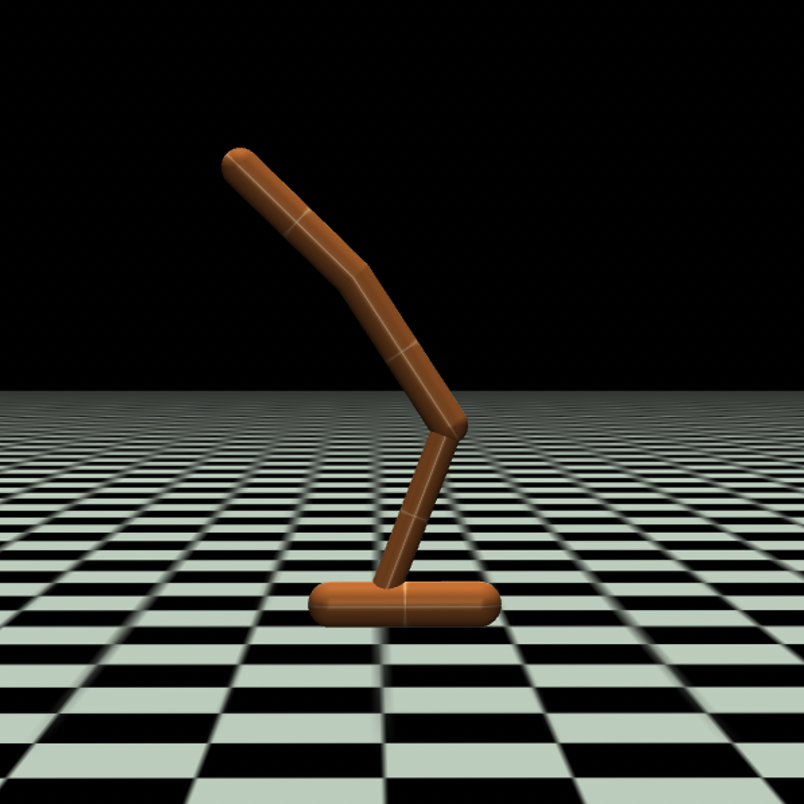}
     \end{subfigure}
     \hspace{0.05cm}
     \begin{subfigure}[b]{0.3\textwidth}
         \centering
         \includegraphics[width=\textwidth]{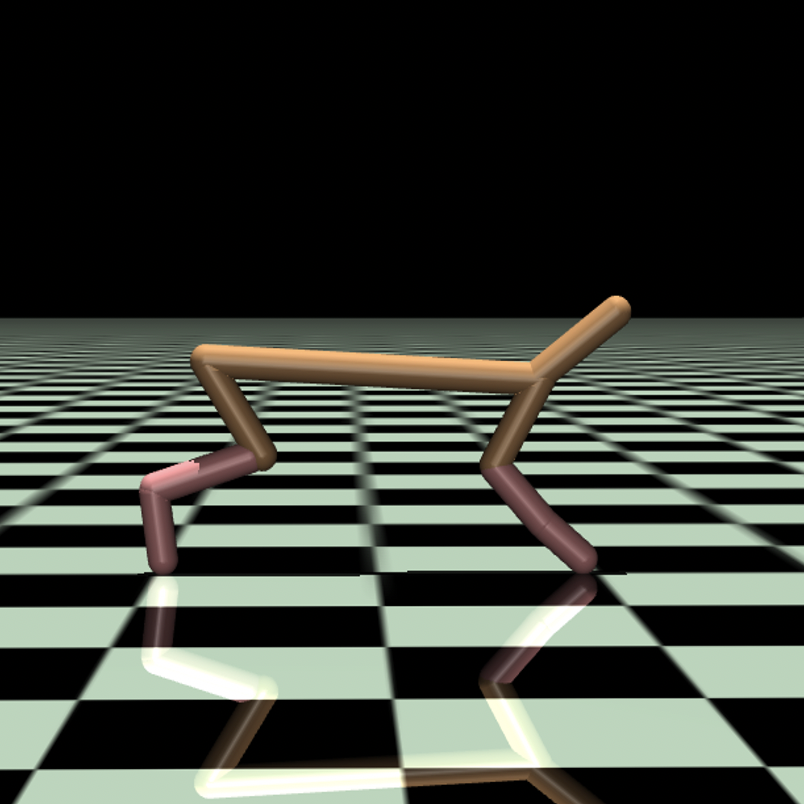}
     \end{subfigure}
          \hspace{0.05cm}
     \begin{subfigure}[b]{0.3\textwidth}
         \centering
         \includegraphics[width=\textwidth]{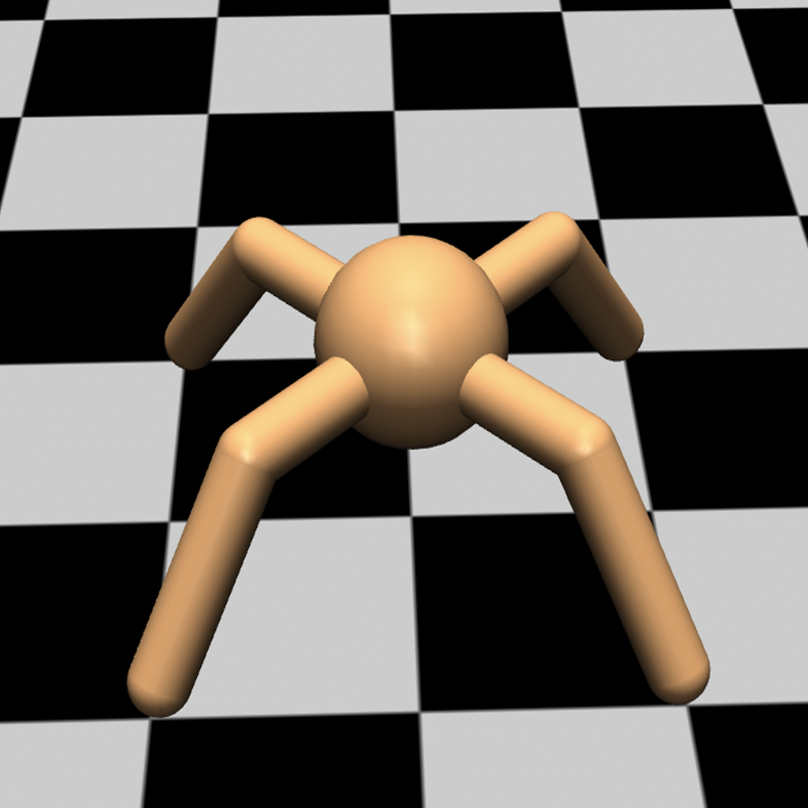}
     \end{subfigure} 
\end{subfigure}
        \caption{From left to right, the Hopper, Half-Cheetah, and Ant environments we use to validate our approaches.}
        \label{fig:mujoco_envs}
\end{figure}

We experiment with the Hopper, Ant, and Half-Cheetah continuous control environments shown in Fig.~\ref{fig:mujoco_envs}. To generate the validation model mismatch, we predefine ranges of mass and friction coefficients as follows: for Hopper, mass $\in [0.7, 1.3]$ and friction $\in [0.7, 1.3]$; Half Cheetah and Ant, mass $\in [0.5, 1.5]$ and friction $\in [0.1, 0.9]$. We scale the friction of every Mujoco geom and the mass of the torso with the same (respective) coefficients. We compare the robustness of agents trained via RAP against: 1) agents trained against a single adversary in a zero-sum game, 2) agents trained using domain randomization, and 3) an agent trained only using PPO and no perturbation mechanism. To train the domain randomization oracle, at each rollout we uniformly sample a friction and mass coefficient from the validation set ranges. We then scale the friction of all geoms and the mass of the torso by their respective coefficients; this constitutes directly training on the validation set which creates a strong baseline. To generate the test set of model mismatch, we take both the highest and lowest friction coefficients from the validation range and apply them to different combinations of individual geoms. For the exact selected combinations, please refer to Appendix~\ref{sec:appendix_holdout}.

\subsection{Computational Details and Reproducibility}
All of our experiments are run on c4.8xlarge 36 vCPU instances on AWS EC2. Our full paper can be reproduced for a cost of $\approx \$100$ (full breakdown in Appendix) but we provide all of the trained policies, data, and code at \url{https://github.com/eugenevinitsky/robust_RL_multi_adversary} to simplify reproducibility. For our RL algorithms we use the RLlib 0.8.0~\cite{liang2017rllib} implementation of PPO~\cite{schulman2017proximal}. For exact hyperparameters, please refer to the Appendix. Since both gradient computations and forwards passes can be batched across the adversaries, there is no additional run-time cost relative to using a single adversary.

\section{Results}\label{sec:mujoco_results}

\begin{figure}
     \centering
     \begin{subfigure}[b]{0.4\textwidth}
         \centering
         \includegraphics[width=\textwidth]{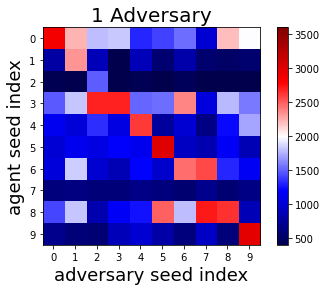}
     \end{subfigure}
     \begin{subfigure}[b]{0.4\textwidth}
         \centering
         \includegraphics[width=\textwidth]{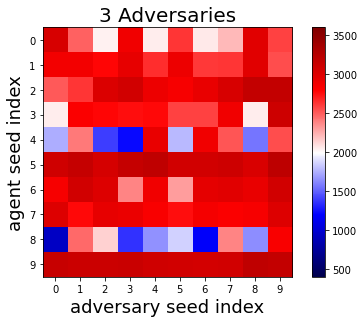}
     \end{subfigure} 
        \caption{Average cumulative reward under swaps for one adversary training (left) and three-adversary training (right). Each square corresponds to 20 trials. In the three adversary case, each square is the average performance against the adversaries from that seed.} 
        \label{fig:overfitting}
\end{figure}

\textbf{\ref{H1} Analysis of Overfitting}\\
A globally minimax optimal adversary should be unexploitable and have a lower bound on performance against any adversary in the adversary class. We investigate the optimality of our policy by asking whether the minimax agent is robust to swaps of adversaries from different training runs, i.e. different seeds.  Fig.~\ref{fig:overfitting} shows the result of these swaps for the one adversary and three adversary case. The diagonal corresponds to playing against the adversaries the agent was trained with while every other square corresponds to playing against adversaries from a different seed. To simplify presentation, in the three adversary case, each square is the average performance against all the adversaries from that seed.

We observe that the agent trained against three adversaries is robust under swaps while the single adversary case is not. For the single adversary case, the mean performance of the agent in each seed is high against its own set of adversaries (the diagonal). This corresponds to the mean reward that would be reported at the end of training. Looking at just the reward is deceptive, the agent is still highly exploitable, as can be seen by its extremely sub-par performance against an adversary from any other seed. Since the adversaries off-diagonal are feasible adversaries, this suggests that we have found a poor local optimum of the objective.

In contrast, the three adversary case is generally robust regardless of which adversary it plays against, suggesting that the use of additional adversaries has made the agent more robust. Of course, it's possible that the adversaries are simply weaker, but as we discuss in Sec.~\ref{H2}, the improved performance on transfer tasks suggests that the robustness across seed swaps is indicative of a genuine improvement in robustness.

\textbf{\ref{H2} Adversary Population Performance}\\
Here we present the results from the validation and holdout test sets described in Section \ref{sec:exp_setup}. We compare the performance of training with adversary populations of size three and five against vanilla PPO, the domain randomization oracle, and the single minimax adversary. 

\begin{figure}
     \centering
     \begin{subfigure}[b]{0.4\textwidth}
         \centering
         \includegraphics[width=\textwidth]{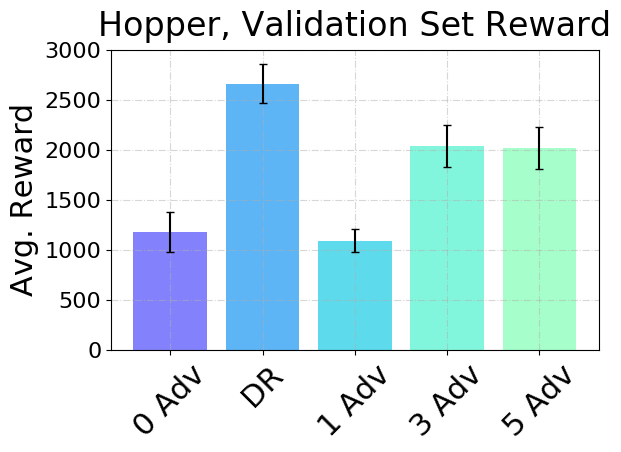}
     \end{subfigure}
          \begin{subfigure}[b]{0.4\textwidth}
         \centering
         \includegraphics[width=\textwidth]{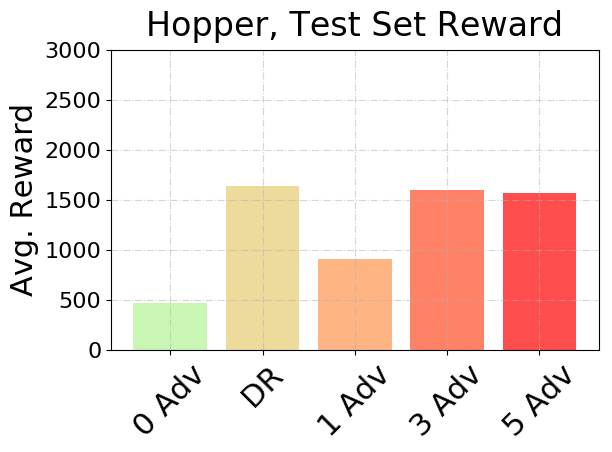}
     \end{subfigure} \\
     \begin{subfigure}[b]{0.4\textwidth}
         \centering
         \includegraphics[width=\textwidth]{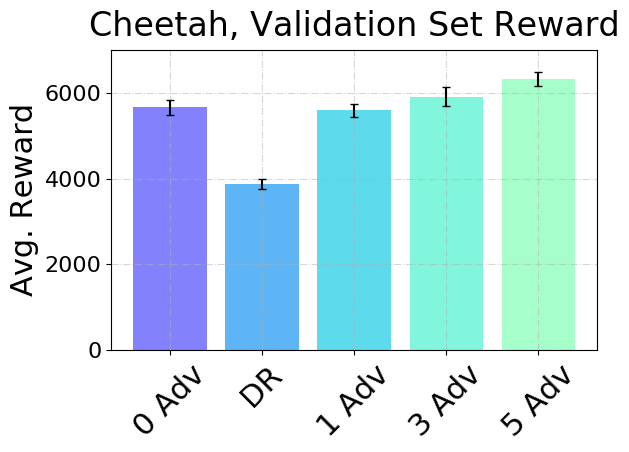}
     \end{subfigure}
     \begin{subfigure}[b]{0.4\textwidth}
         \centering
         \includegraphics[width=\textwidth]{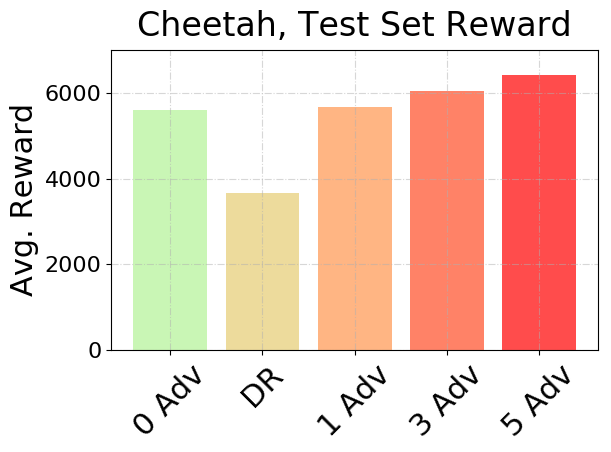}
     \end{subfigure} \\
     \begin{subfigure}[b]{0.4\textwidth}
         \centering
         \includegraphics[width=\textwidth]{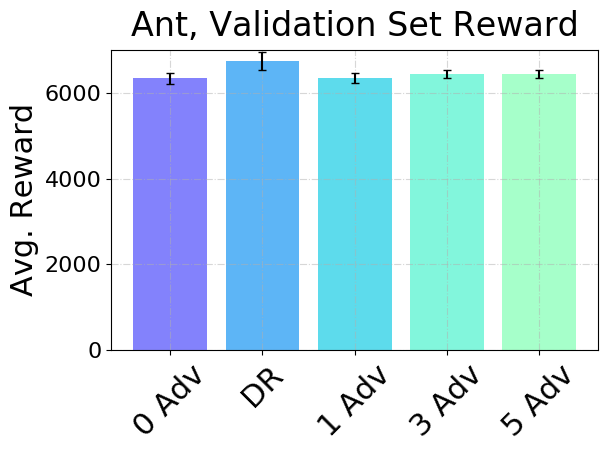}
     \end{subfigure}
    \begin{subfigure}[b]{0.4\textwidth}
         \centering
         \includegraphics[width=\textwidth]{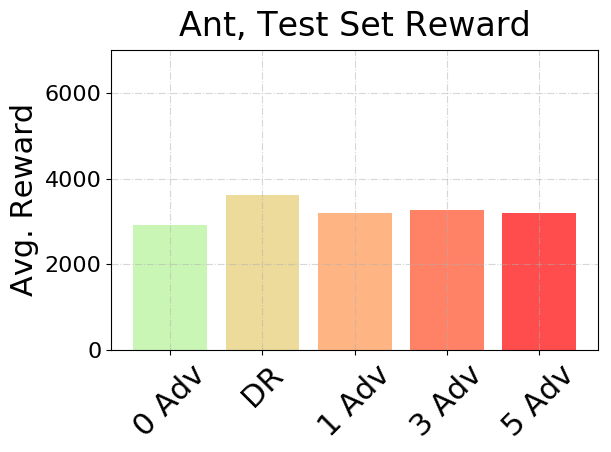}
     \end{subfigure}
        \caption{Average reward for Ant, Hopper, and Cheetah environments across ten seeds and across the validation set (left column) and across the holdout test set (right column). We compare vanilla PPO, the domain randomization oracle, and the minimax adversary against RAP of size three and five. Bars represent the mean and the arms represent the std. deviation. Both are computed over 20 rollouts for each test-set sample. The std. deviation for the test set are not reported here for visual clarity due to the large variation in holdout test difficulty.}
        \label{fig:mujoco_robustness}
\end{figure}

Fig.\ref{fig:mujoco_robustness} shows the average reward (the average of ten seeds across the validation or test sets respectively) for each environment. Table \ref{tab:mujoco_results} gives the corresponding numerical values and the percent change of each policy from the baseline. Standard deviations are omitted on the test set due to wide variation in task difficulty; the individual tests that we aggregate here are reported in the Appendix Sec.~\ref{sec:appendix_results} with appropriate error bars. In all environments we achieve a higher reward across both the validation and holdout test set using RAP of size three and/or five when compared to the single minimax adversary case. These results from testing on new environments with altered dynamics supports hypothesis \ref{H1} that training with a population of adversaries leads to more robust policies than training with a single adversary.

\begin{figure}
\begin{subfigure}[b]{\textwidth}
     \centering
     \begin{subfigure}[b]{0.4\textwidth}
         \centering
         \includegraphics[width=\textwidth]{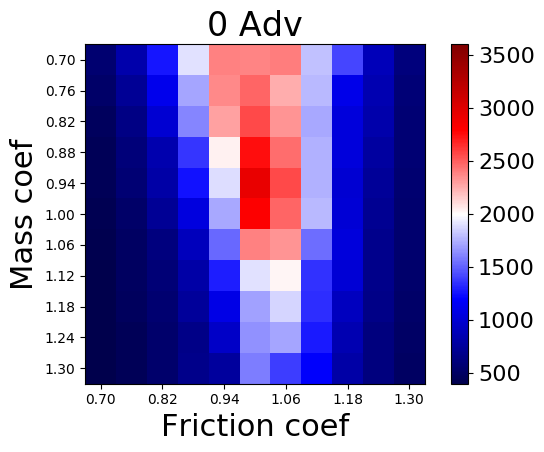}
     \end{subfigure}
     \begin{subfigure}[b]{0.4\textwidth}
         \centering
         \includegraphics[width=\textwidth]{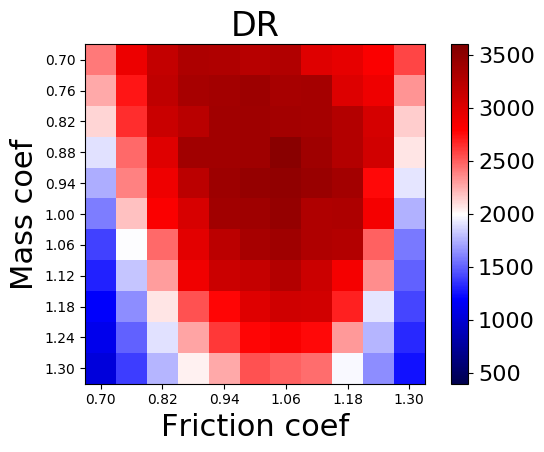}
     \end{subfigure}
     \begin{subfigure}[b]{0.4\textwidth}
         \centering
         \includegraphics[width=\textwidth]{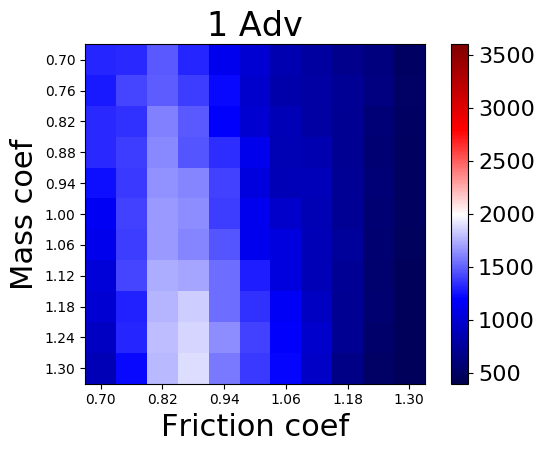}
     \end{subfigure} 
     \begin{subfigure}[b]{0.4\textwidth}
         \centering
         \includegraphics[width=\textwidth]{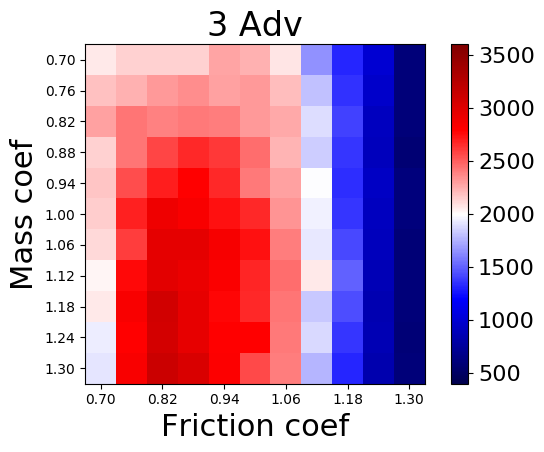}
     \end{subfigure}
        \caption{Hopper.}
        \label{fig:hopheat}
\end{subfigure}

\begin{subfigure}[b]{\textwidth}
\centering
     \begin{subfigure}[b]{0.4\textwidth}
         \centering
         \includegraphics[width=\textwidth]{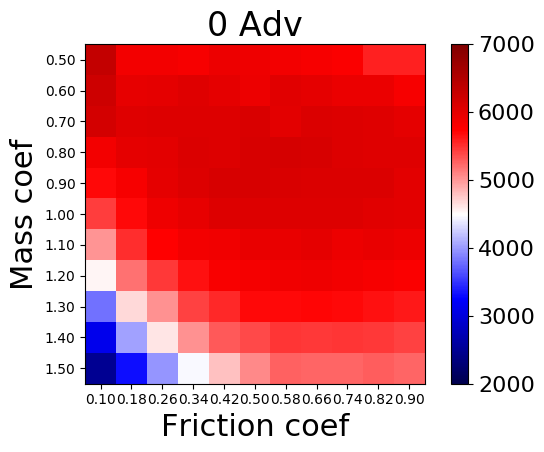}
     \end{subfigure}
     \begin{subfigure}[b]{0.4\textwidth}
         \centering
         \includegraphics[width=\textwidth]{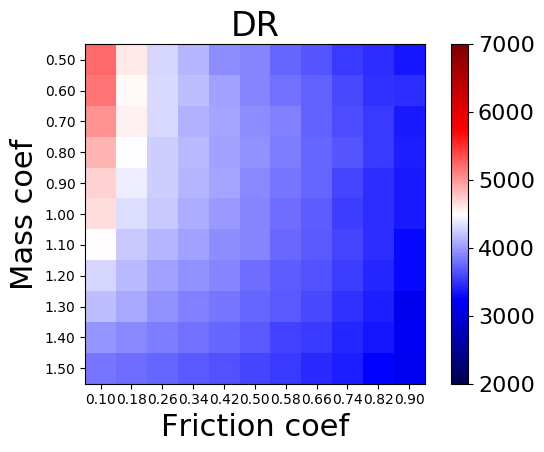}
     \end{subfigure}
     \begin{subfigure}[b]{0.4\textwidth}
         \centering
         \includegraphics[width=\textwidth]{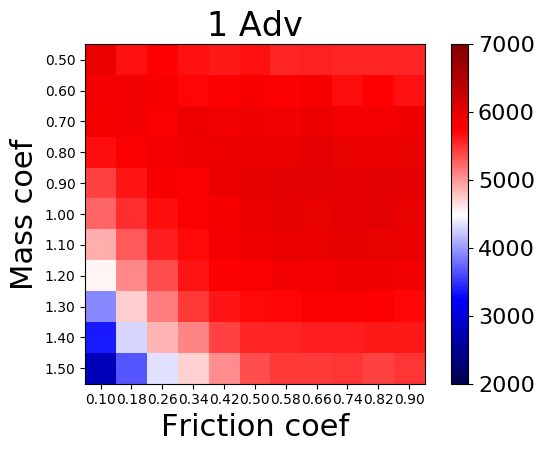}
     \end{subfigure} 
     \begin{subfigure}[b]{0.4\textwidth}
         \centering
         \includegraphics[width=\textwidth]{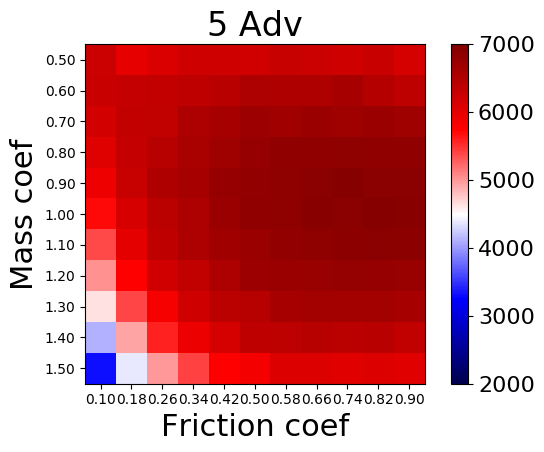}
     \end{subfigure}
     \caption{Half Cheetah.}
        \label{fig:hcheat}
     \end{subfigure}
        \caption{Average reward across ten seeds on each validation set parametrization -- friction coefficient on the x-axis and mass coefficient on the y-axis. Going from left to right: Row 1 - 0 Adversary and Domain Randomization (Hopper), Row 2 - 1 Adversary and 3 Adversaries (Hopper), Row 3 - 0 Adversary and Domain Randomization (Cheetah), Row 4 - 1 Adversary and 5 Adversaries (Cheetah).}
        \label{fig:heatmaps}
\end{figure}

For a more detailed comparison of robustness across the validation set,
Fig. \ref{fig:heatmaps} shows heatmaps of the performance across all the mass, friction coefficient combinations. Here we highlight the heatmaps for Hopper and Half Cheetah for vanilla PPO, domain randomization, single adversary, and best adversary population size. Additional heatmaps for other adversary population sizes and the Ant environment can be found in Appendix Sec.~\ref{sec:appendix_results}. Note that Fig. \ref{fig:hopheat} is an example of a case where a single adversary has negligible effect on or slightly reduces the performance of the resultant policy on the validation set. This supports our hypothesis that a single adversary can actually lower the robustness of an agent.
This result is in contrast to those observed in ~\cite{pinto2017robust}; we conjecture that their hand-designed parametrization of the adversary (forces applied to carefully selected leg joints) may be the cause of the difference.

\begin{table}
    \centering
    \small

        \begin{tabular}{c||c|c|c|c|c||c|c|c|c|c}
    & \multicolumn{5}{c||}{Validation} &  \multicolumn{5}{c}{Test}  \\\hline
       Hopper   & 0 Adv & DR & 1 Adv & 3 Adv & 5 Adv & 0 Adv & DR & 1 Adv & 3 Adv & 5 Adv\\ \hline
        Mean Rew. & 1182&	2662&	1094&	\textbf{2039}&	2021 &472&	1636&	913&	\textbf{1598} &	1565 \\
        \% Change& &125&	-7.4&\textbf{	72.6}&	71& & 246&	93.4&	\textbf{238}&	231 \\\hline 
    \end{tabular}
        \begin{tabular}{c||c|c|c|c|c||c|c|c|c|c}
    & \multicolumn{5}{c||}{Validation} &  \multicolumn{5}{c}{Test}  \\\hline
        Cheetah   & 0 Adv & DR & 1 Adv & 3 Adv & 5 Adv & 0 Adv & DR & 1 Adv & 3 Adv & 5 Adv\\ \hline
        Mean Rew. & 5659&	3864&	5593&	5912&	\textbf{6323}& 5592&	3656&	5664&	6046&	\textbf{6406} \\
        \% Change && -32&	-1.2&	4.5&	\textbf{11.7}& &-35&	1.3&	8.1&	\textbf{14.6} \\\hline 
    \end{tabular}
        \begin{tabular}{c||c|c|c|c|c||c|c|c|c|c}
    & \multicolumn{5}{c||}{Validation} &  \multicolumn{5}{c}{Test}  \\\hline
       Ant   & 0 Adv & DR & 1 Adv & 3 Adv & 5 Adv &0 Adv & DR & 1 Adv & 3 Adv & 5 Adv\\ \hline
        Mean Rew. & 6336 &6743&	6349	&6432	&\textbf{6438} &2908	&3613&	3206&	\textbf{3272}&	3203 \\
        \% Change& &6.4&	0.2&	1.5&	\textbf{1.6}& &24.3&	10.2&	\textbf{12.5}&	10.2 \\\hline 
    \end{tabular}
    \caption{Average reward and \% change from vanilla PPO (0 Adv) for Hopper, Cheetah, and Ant environments across ten seeds and across the validation (left) or holdout test set (right). Across all environments, we see consistently higher robustness using RAP than the minimax adversary. Most robust adversarial approach is bolded.}
    \label{tab:mujoco_results}
\end{table}

\textbf{\ref{H3} Effect of Domain Randomization Parametrization}\\
\label{sec:dr_param}
From Fig. \ref{fig:mujoco_robustness}, we see that in the Ant and Hopper domains, the oracle achieves the highest transfer reward in the validation set as expected since the oracle is trained directly on the validation set. Interestingly, we found that the domain randomization policy performed much worse on the Half Cheetah environment, despite having access to the mass and friction coefficients during training. Looking at the performance for each mass and friction combination in Fig.~\ref{fig:hcheat}, we found that the DR agent was able to perform much better at the low friction coefficients and learned to prioritize those values at the cost of significantly worse performance on average. This highlights a potential issue with domain randomization: while training across a wide variety of dynamics parameters can increase robustness, naive parametrizations can cause the policy to exploit subsets of the randomized domain and lead to a brittle policy. 

\begin{figure}
\begin{subfigure}[b]{\textwidth}
\centering
     \begin{subfigure}[b]{0.4\textwidth}
         \centering
         \includegraphics[width=\textwidth]{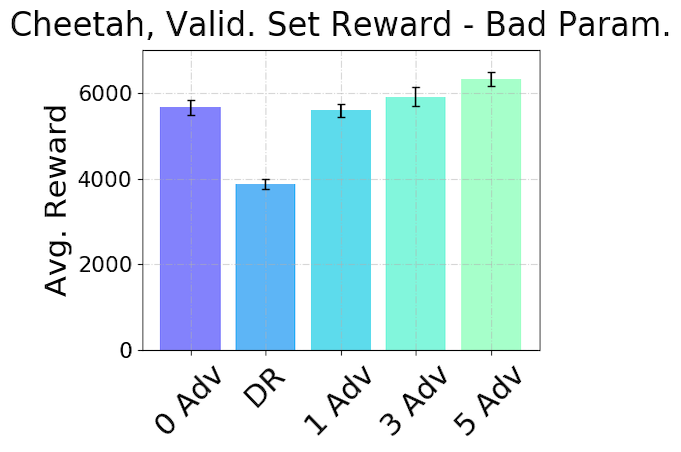}
     \end{subfigure} 
     \begin{subfigure}[b]{0.4\textwidth}
         \centering
         \includegraphics[width=\textwidth]{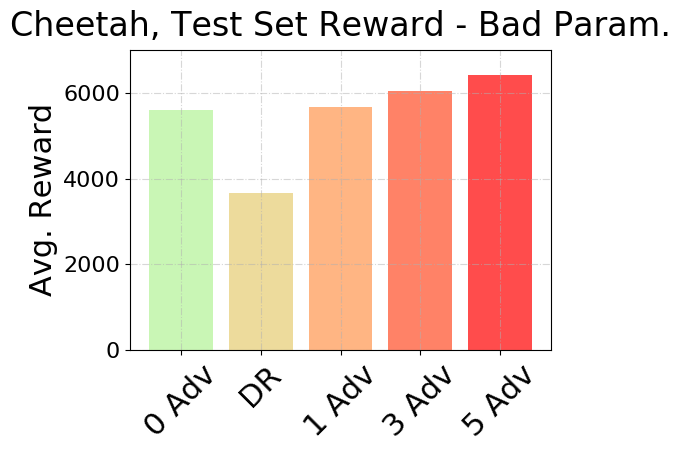}
     \end{subfigure} 
     \caption{Bad Parametrization.}
\end{subfigure} \\
\begin{subfigure}[b]{\textwidth}
    \centering
     \begin{subfigure}[b]{0.4\textwidth}
         \centering
         \includegraphics[width=\textwidth]{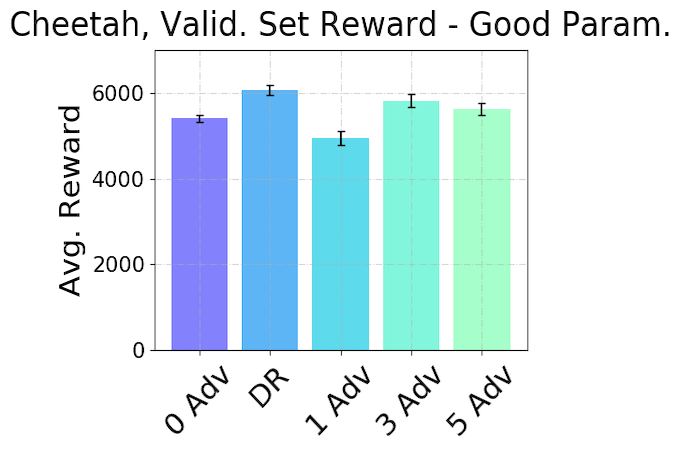}
     \end{subfigure}
     \begin{subfigure}[b]{0.4\textwidth}
         \centering
         \includegraphics[width=\textwidth]{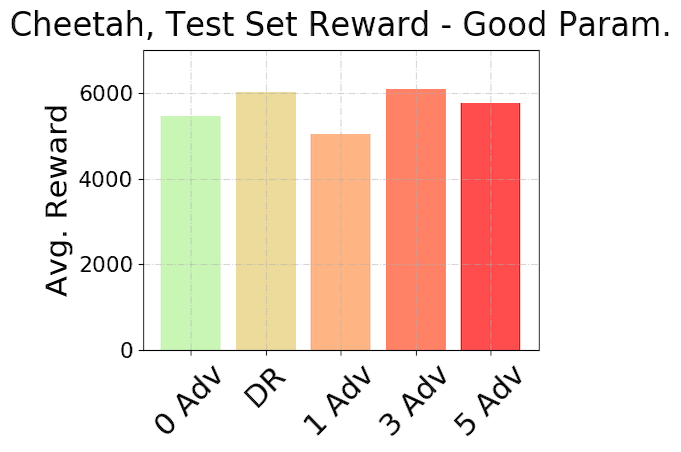}
     \end{subfigure}
          \caption{Good Parametrization.}

\end{subfigure}
        \caption{Average reward for Half Cheetah environment across ten seeds. (a) shows the average reward when trained with a `bad' friction parametrization which lead to DR not learning a robust agent policy, and (b) shows the average reward when trained with a `good' friction parametrization.}
        \label{fig:hc_goodbad}
\end{figure}

\begin{figure}
     \centering
     \begin{subfigure}[b]{0.4\textwidth}
         \centering
         \includegraphics[width=\textwidth]{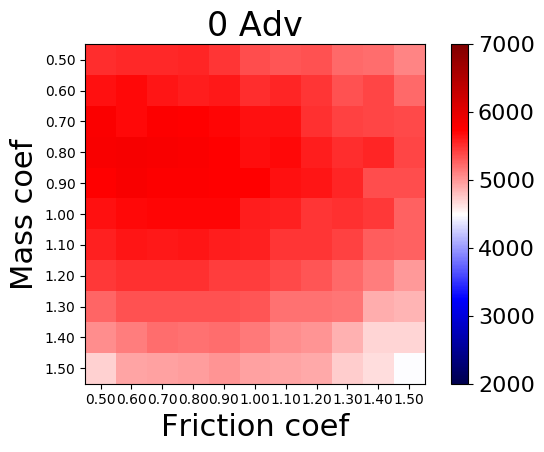}
     \end{subfigure}
     \begin{subfigure}[b]{0.4\textwidth}
         \centering
         \includegraphics[width=\textwidth]{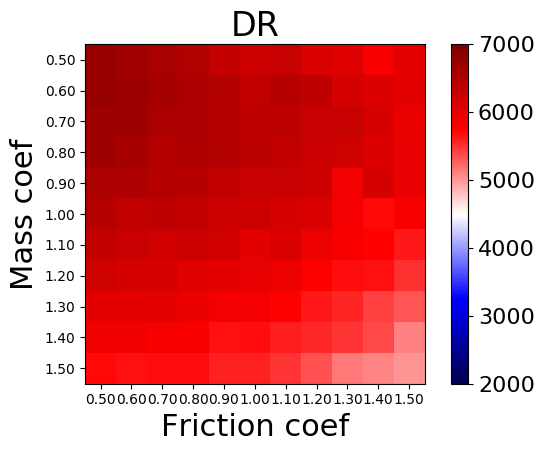}
     \end{subfigure}
     \begin{subfigure}[b]{0.4\textwidth}
         \centering
         \includegraphics[width=\textwidth]{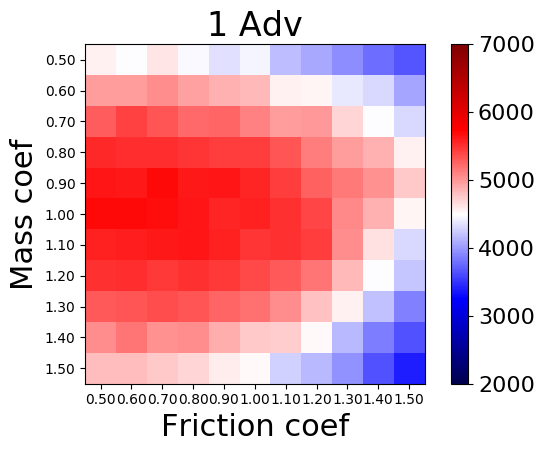}
     \end{subfigure} 
     \begin{subfigure}[b]{0.4\textwidth}
         \centering
         \includegraphics[width=\textwidth]{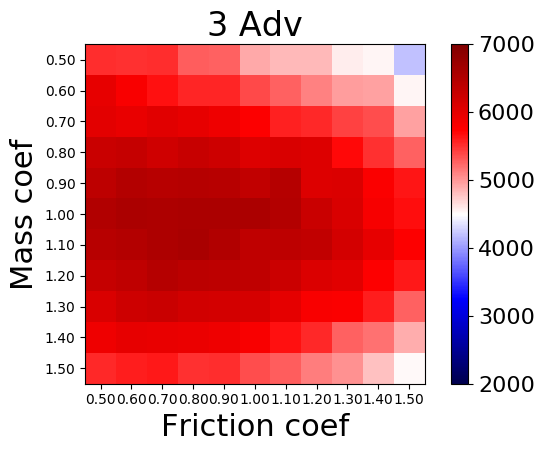}
     \end{subfigure}
        \caption{Half Cheetah Heatmap (Good parametrization) -- friction coefficient on the x-axis and mass coefficient on the y-axis. Going from left to right: Row 1 - 0 adversary and Domain Randomization, Row 2 - 1 Adversary and 3 Adversaries.}
        \label{fig:hcheat_good}
\end{figure}

We hypothesize that this is due to the DR objective in Eq. \ref{eq:domain_randomization} optimizing in expectation over the sampling range. To test this, we created a separate range of `good' friction parameters $[0.5, 1.5]$ and compared the robustness of a DR policy trained with a `good` range against a DR policy trained with a `bad' range $[0.1, 0.9]$ in Fig. \ref{fig:hc_goodbad}. Here we see that a `good' parametrization leads to the expected result where domain randomization is the most robust. We observe that domain randomization, under the `bad' paramtrization underperforms adversarial training on the validation set despite the validation set literally constituting the training set for domain randomization. This suggests that underlying optimization difficulties are partially to blame for the poor performance of domain randomization. Notably, the adversary-based methods are not susceptible to the same parametrization issues. 

Prior work, EPOpt~\cite{rajeswaran2016epopt}, has addressed this issue by replacing the uniform average across distributions with the conditional value at risk (CVaR)~\cite{chow2015risk, tamar2015optimizing}, a soft version of the minimax objective where the optimization is only performed over a small percentage of the worst performing parameters. This interesting approach to align the domain randomization objective with the minimax objective could be made compatible with our approach by training using a subset of the strongest adversaries. 

\begin{figure}
     \centering
     \begin{subfigure}[b]{0.4\textwidth}
         \centering
         \includegraphics[width=\textwidth]{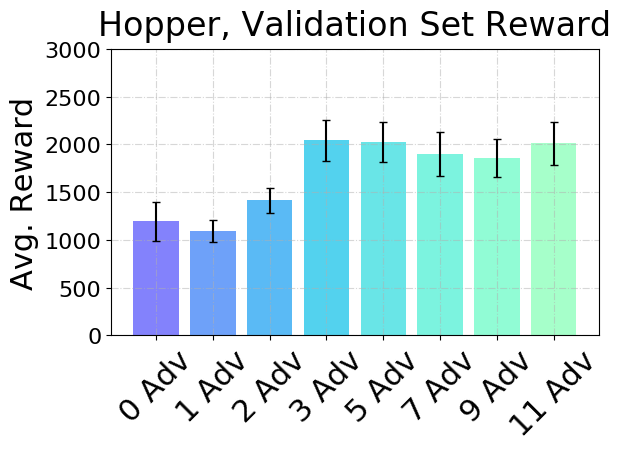}
     \end{subfigure}
     \begin{subfigure}[b]{0.4\textwidth}
         \centering
     \includegraphics[width=\textwidth]{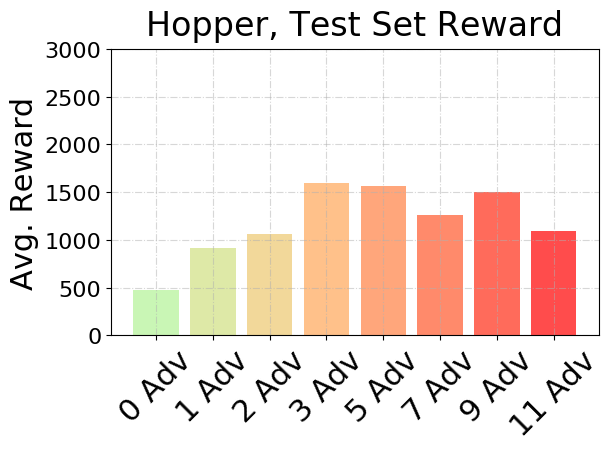}
     \end{subfigure} 
        \caption{Average reward for Hopper across varying adversary number. Due to the wide variation in test set difficulty, standard deviations are not depicted here.}
        \label{fig:adv_number_sweep}
\end{figure}

\textbf{\ref{H4} Increasing Adversary Population Size}\\
We investigate whether \textbf{RAP} is robust to adversary number as this would be a useful property to minimize hyperparameter search.
 Here we hypothesize that while having more adversaries can represent a wider range of dynamics to learn to be robust to, we expect there to be diminishing returns due to the decreased batch size that each adversary receives (total number of environment steps is held constant across all training variations). We expect decreasing batch size to lead to worse agent policies since the batch will contain under-trained adversary policies that the agent will learn to exploit. We cap the number of adversaries at eleven as our machines ran out of memory at this value. We run ten seeds for every adversary value and Fig.~\ref{fig:adv_number_sweep} shows the results for Hopper. Agent robustness on the test set increases monotonically up to three adversaries and roughly begins to decrease after that point. This suggests that a trade-off between adversary number and performance exists although we do not definitively show that diminishing batch sizes is the source of this trade-off. However, we observe in Fig.~\ref{fig:mujoco_robustness} that both three and five adversaries perform well across all studied Mujoco domains.

\section{Conclusions and Future Work}
In this work we demonstrate that the use of a single adversary to approximate the solution to a minimax problem does not consistently lead to improved robustness. We propose a solution through the use of multiple adversaries (RAP), and demonstrate that this provides robustness across a variety of robotics benchmarks. We also compare RAP with domain randomization and demonstrate that while DR can lead to a more robust policy, it requires careful parametrization of the domain we sample from to ensure robustness. RAP does not require this tuning, allowing for use in domains where appropriate tuning requires extensive prior knowledge or expertise.

There are several open questions stemming from this work. While we empirically demonstrate the effects of RAP, we do not have a compelling theoretical understanding of why multiple adversaries are helping. Perhaps RAP helps approximate a mixed Nash equilibrium as discussed in Sec.~\ref{sec:introduction} or perhaps population based training increases the likelihood that one of the adversaries is strong? Would the benefits of RAP disappear if a single adversary had the ability to represent mixed Nash (for example, by adding a source of randomness to the adversary state)? Another interesting question to ask is whether the minimax games described here satisfy the "games-of-skill" hypothesis~\cite{czarnecki2020real} which would provide an optimization-based reason for including adversary populations.

There are some interesting extensions of this work that we would like to pursue. We have looked at the robustness of our approach in simulated settings; future work will examine whether this robustness transfers to real-world settings. Additionally, our agents are currently memory-less and therefore cannot perform adversary identification; it would be worthwhile to see if the auxiliary task of adversary identification leads to a robust system-identification procedure that improves transfer performance. Our adversaries can also be viewed as forming a task distribution, allowing them to be used in continual learning approaches like MAML~\cite{nagabandi2018deep} where domain randomization is frequently used to construct task distributions. 

Finally, here we apply adversary populations to the noisy robust MDP; applying the adversary action to the agent action represents a restricted class of dynamical systems. The transfer tests used in this work may not even be included in the set of dynamical systems that can be represented by this adversary class; this restriction may be reducing the transfer performance of the minimax approach. In future work we would like to consider a wider range of dynamical systems by using a more powerful adversary class that can control the dynamics directly.

\section*{Acknowledgments}
The authors would like to thank Lerrel Pinto for help understanding and reproducing "Robust Adversarial Reinforcement Learning" as well as insightful discussions of our problem. Additionally, we would like to thank Natasha Jaques and Michael Dennis who helped us develop intuition for what the single adversary case might be flawed. Eugene Vinitsky is a recipient of an NSF Graduate Research Fellowship and funded by National Science Foundation under Grant Number CNS-1837244. Yuqing Du is funded by a Berkeley AI Research fellowship \& ONR through PECASE N000141612723. Computational resources for this work were provided by an AWS Machine Learning Research grant. This material is also based upon work supported by the U.S. Department of Energy’s Office of Energy Efficiency and Renewable Energy (EERE) award number CID DE-EE0008872. The views expressed herein do not necessarily represent the views of the U.S. Department of Energy or the United States Government.

\bibliography{arxiv_version}
\bibliographystyle{abbrv}

\clearpage
\appendix
\appendixpage
\addappheadtotoc
\section{Full Description of the MDPs}
\label{sec: mujoco-describe}

We use the Mujoco ant, cheetah, and hopper environments as a test of the efficacy of our strategy versus the 0 adversary, 1 adversary, and domain randomization baselines. We use the Noisy Action Robust MDP formulation~\cite{tessler2019action} for our adversary parametrization. 
If the normal system dynamics are 
$$s_{k+1} = s_k + f(s_k, a_k) \Delta t $$ 
the system dynamics under the adversary are 
$$s_{k+1} = s_k + f(s_k, a_k + a^{\text{adv}}_k) \Delta t $$ 

where $a^{\text{adv}}_k$ is the adversary action at time k.

The notion here is that the adversary action is passed through the dynamics function and represents some additional set of dynamics. It is standard to clip actions within some boundary but clipping the sum would allow the agent to "cancel" the adversary by always keeping its action at the bounds of the action space. Since we want the adversary to always affect the dynamics irrespective of agent action, 
we clip the agent and adversary actions separately. The agent is clipped between $[-1, 1]$ in all environments and the adversary is clipped between $[-.25, .25]$.

The MDP through which we train the agent policy is characterized by the following states, actions, and rewards:
\begin{itemize}
    \item $s^{\text{agent}}_t = \left[o_t, a_{t-1}\right] $  where $o_t$ is an observation returned by the environment, and $a_t$ is the action taken by the agent.
    \item We use the standard rewards provided by the OpenAI Gym Mujoco environments at  \url{https://github.com/openai/gym/tree/master/gym/envs/mujoco}. For the exact functions, please refer to the code at \url{https://github.com/eugenevinitsky/robust_RL_multi_adversary}.
    \item $a^{\text{agent}}_t \in \left[a_\text{min}, a_\text{max} \right]^n$.
\end{itemize}

The MDP for adversary $i$ is the following:
\begin{itemize}
    \item $s_t = s^{\text{agent}}_t$. The adversary sees the same states as the agent.
    \item The adversary reward is the negative of the agent reward.
    \item $a^{\text{adv}}_t \in \left[a^{\text{adv}}_\text{min}, a^{\text{adv}}_\text{max} \right]^n$.
\end{itemize}

For our domain randomization Hopper baseline, we use the following randomization: at each rollout, we scale the friction of all joints by a single value uniformly sampled from [0.7, 1.3]. We also randomly scale the mass of the 'torso' link by a single value sampled from [0.7, 1.3]. For Half-Cheetah and Ant the range for friction is [0.1, 0.9] and for mass the range is [0.5, 1.5].

\section{Holdout Tests}
\label{sec:appendix_holdout}
In this section we describe in detail all of the holdout tests used.

\subsection{Hopper}

\begin{figure}
    \centering
    \includegraphics[scale=0.5]{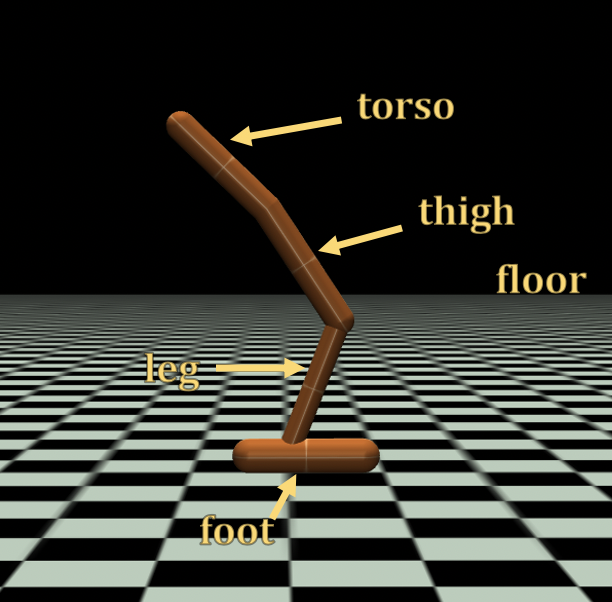}
    \caption{Labelled Body Segments of Hopper}
    \label{fig:hopperlabelled}
\end{figure}

\begin{table}
    \centering
       \captionsetup{justification=centering, skip=5pt}
    \caption{Hopper Holdout Test Descriptions}
    \begin{tabular}{c||c|c}
        \hline
       Test  & Body with Friction Coeff 1.3 & Body with Friction Coeff 0.7 \\ \hline
        A & Torso, Leg & Floor, Thigh, Foot \\
        B & Floor, Thigh & Torso, Leg, Foot \\
        C & Foot, Leg & Floor, Torso, Thigh\\
        D & Torso, Thigh, Floor & Foot, Leg \\
        E & Torso, Foot & Floor, Thigh, Leg \\
        F & Floor, Thigh, Leg & Torso, Foot \\
        G & Floor, Foot & Torso, Thigh, Leg \\
        H & Thigh, Leg & Floor, Torso, Foot \\ \hline
    \end{tabular} \\
    \label{tab:hopper_holdout}
\end{table}

The Mujoco geom properties that we modified are attached to a particular body and determine its appearance and collision properties. For the Mujoco holdout transfer tests we pick a subset of the hopper `geom' elements and scale the contact friction values by maximum friction coefficient, $1.3$. Likewise, for the rest of the `geom' elements, we scale the contact friction by the minimum value of $0.7$. The body geoms and their names are visible in Fig.~\ref{fig:hopperlabelled}.

The exact combinations and the corresponding test name are indicated in Table \ref{tab:hopper_holdout} for Hopper.

\subsection{Cheetah}

\begin{figure}
    \centering
    \includegraphics[scale=0.5]{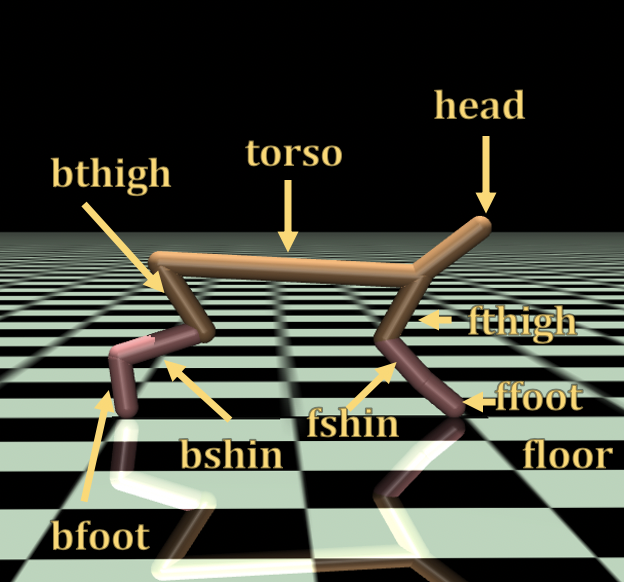}
    \caption{Labelled Body Segments of Cheetah}
    \label{fig:cheetahlabelled}
\end{figure}

\begin{table}
    \centering
       \captionsetup{justification=centering, skip=5pt}
    \caption{Cheetah Holdout Test Descriptions. Joints in the table receive the maximum friction coefficient of 0.9. Joints not indicated have friction coefficient 0.1}
    \begin{tabular}{c||c}
        \hline
       Test  & Geom with Friction Coeff 0.9 \\ \hline
        A & Torso, Head, Fthigh \\
        B & Floor, Head, Fshin \\
        C & Bthigh, Bshin, Bfoot \\
        D & Floor, Torso, Head\\
        E & Floor, Bshin, Ffoot\\
        F & Bthigh, Bfoot, Ffoot \\
        G & Bthigh, Fthigh, Fshin \\
        H & Head, Fshin, Ffoot \\ \hline
    \end{tabular} \\
    \label{tab:cheetahholdout}
\end{table}

The Mujoco geom properties that we modified are attached to a particular body and determine its appearance and collision properties. For the Mujoco holdout transfer tests we pick a subset of the cheetah `geom' elements and scale the contact friction values by maximum friction coefficient, $0.9$. Likewise, for the rest of the `geom' elements, we scale the contact friction by the minimum value of $0.1$. The body geoms and their names are visible in Fig.~\ref{fig:cheetahlabelled}.

The exact combinations and the corresponding test name are indicated in Table \ref{tab:cheetahholdout} for Hopper.

\subsection{Ant}

\begin{figure}
    \centering
    \includegraphics[scale=0.5]{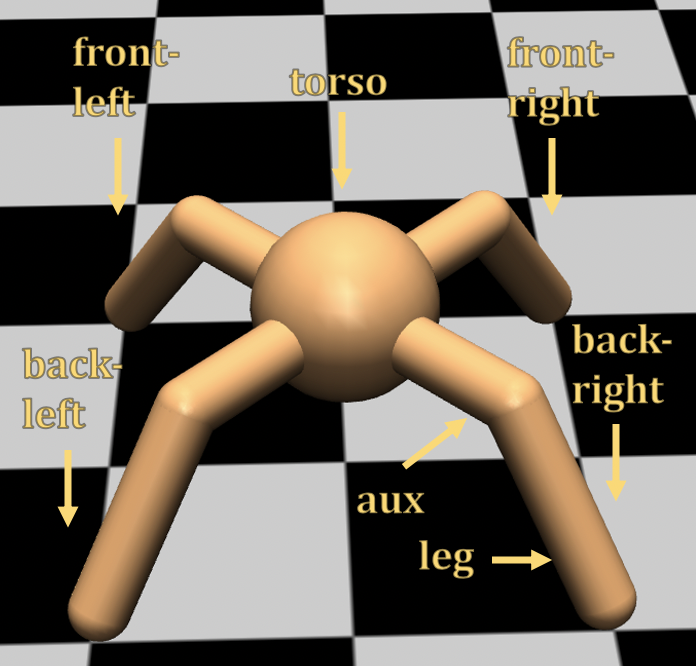}
    \caption{Labelled Body Segments of Ant}
    \label{fig:antlabelled}
\end{figure}

\begin{table}
    \centering
       \captionsetup{justification=centering, skip=5pt}
    \caption{Ant Holdout Test Descriptions. Joints in the table receive the maximum friction coefficient of 0.9. Joints not indicated have friction coefficient 0.1}
    \begin{tabular}{c||c}
        \hline
       Test  & Geom with Friction Coeff 0.9 \\ \hline
        A & Front-Leg-Left, Aux-Front-Left, Aux-Back-Left \\
        B & Torso, Aux-Front-Left, Back-Leg-Right \\
        C & Front-Leg-Right, Aux-Front-Right, Back-Leg-Left \\
        D & Torso, Front-Leg-Left, Aux-Front-Left\\
        E & Front-Leg-Left, Aux-Front-Right, Aux-Back-Right\\
        F & Front-Leg-Right, Back-Leg-Left, Aux-Back-Right \\
        G & Front-Leg-Left, Aux-Back-Left, Back-Leg-Right \\
        H & Aux-Front-Left, Back-Leg-Right, Aux-Back-Right \\ \hline
    \end{tabular} \\
    \label{tab:antholdout}
\end{table}

We will use torso to indicate the head piece, leg to refer to one of the four legs that contact the ground, and 'aux' to indicate the geom that connects the leg to the torso. Since the ant is symmetric we adopt a convention that two of the legs are front-left and front-right and two legs are back-left and back-right. Fig.~\ref{fig:antlabelled} depicts the convention.
For the Mujoco holdout transfer tests we pick a subset of the ant `geom' elements and scale the contact friction values by maximum friction coefficient, $0.9$. Likewise, for the rest of the `geom' elements, we scale the contact friction by the minimum value of $0.1$. 

The exact combinations and the corresponding test name are indicated in Table \ref{tab:antholdout} for Hopper.


\section{Results}
\label{sec:appendix_results}
Here we recompute the values of all the results and display them with appropriate standard deviations in tabular form. Tables~\ref{table:mujoco_hopper},~\ref{table:mujoco_cheetah},~\ref{table:mujoco_ant} contain the test results with appropriate standard deviations for Hopper, Half-Cheetah, and Ant respectively.

\begin{table*}
\begin{tabular}
{|p{2cm}||p{2cm}|p{2cm}|p{2cm}|p{2cm}|p{2cm}|}

 \hline
 
Test Name & 0 Adv & 1 Adv & 3 Adv & Five Adv & Domain Rand\\
 \hline
 Test A  & $410 \pm 140$  & $1170 \pm 570$ &   \bm{$2210 \pm 630$} & $2090 \pm 920$ & $1610 \pm 310$\\
 Test B  & $430 \pm 150$   & $1160 \pm 540$   & \bm{$2240 \pm 730$} & $2200 \pm 880$ & $1610 \pm 290$\\
 Test C & $560 \pm 120$  & $490 \pm 150$&  $610 \pm 250$ & $580 \pm 120$ & \bm{$1660 \pm 260$}\\
 Test D & $420 \pm 150$ & $1140 \pm 560$&  \bm{$2220 \pm 680$} & $2130 \pm 890$ & $1612 \pm 360$\\
 Test E & $550 \pm 120$  & $500 \pm 150$ & $600 \pm 240$ & $590 \pm 120$ & \bm{$1680 \pm 280$}\\
 Test F & $420 \pm 150$  & $1200 \pm 620$   & $2080 \pm 750$ & \bm{$2160 \pm 890$} & $1650 \pm 360$\\
 Test H & $560 \pm 130$  & $500 \pm 140$ &$600 \pm 230$ & $600 \pm 140$ & \bm{$1710 \pm 370$}\\
 Test G & $420 \pm 150$  & $1160 \pm 590$ & \bm{$2210 \pm 680$} & $2160 \pm 920$ & $1560 \pm 340$\\
 \hline
\end{tabular}
\caption{Results on holdout tests for each of the tested approaches for Hopper. Bolded values have the highest mean}
\label{table:mujoco_hopper}
\end{table*}

\begin{table*}
\begin{tabular}
{|p{2cm}||p{2cm}|p{2cm}|p{2cm}|p{2cm}|p{2cm}|}

 \hline
 
Test Name & 0 Adv & 1 Adv & 3 Adv & Five Adv & Domain Rand\\
 \hline
 Test A  & $4400 \pm 2160$  & $5110 \pm 730$ &   $4960 \pm 1280$ & \bm{$5560 \pm 1060$} & $2800 \pm 1540$\\
 Test B  & $6020 \pm 880$   & $5980 \pm 290$   & $6440 \pm 1620$ & \bm{$6880 \pm 1090$} & $3340 \pm 600$\\
 Test C & $5880 \pm 1030$  & $5730 \pm 640$&  \bm{$6740 \pm 1190$} & $6410 \pm 790$ & $4280 \pm 240$\\
 Test D & $5990 \pm 940$ & $5960 \pm 260$&  $6430 \pm 1610$ & \bm{$6880 \pm 1090$} & $3360 \pm 570$\\
 Test E & $5570 \pm 570$  & $5670 \pm 290$ & $5800 \pm 1316$ & \bm{$6530 \pm 1250$} & $3720 \pm 540$\\
 Test F & $5870 \pm 750$  & $5800 \pm 350$   & $6500 \pm 1100$ & \bm{$6770 \pm 1070$} & $3810 \pm 330$\\
 Test H & $5310 \pm 1060$  & $5270 \pm 700$ & $5610 \pm 720$ & \bm{$5660 \pm 980$} & $4560 \pm 560$\\
 Test G & $5710 \pm 650$  & $5790 \pm 300$ & $5890 \pm 1240$ & \bm{$6560 \pm 1240$} & $3380 \pm 720$\\
 \hline
\end{tabular}
\caption{Results on holdout tests for each of the tested approaches for Half Cheetah. Bolded values have the highest mean}
\label{table:mujoco_cheetah}
\end{table*}

\begin{table*}[h!]
\begin{tabular}
{|p{2cm}||p{2cm}|p{2cm}|p{2cm}|p{2cm}|p{2cm}|}

\hline
 
Test Name & 0 Adv & 1 Adv & 3 Adv & Five Adv & Domain Rand\\
 \hline
 Test A  & $590 \pm 650$  & $730 \pm 630$ &   $600 \pm 440$ & $560 \pm 580$ & \bm{$900 \pm 580$}\\
 Test B  & $5240 \pm 280$   & $5530 \pm 200$   & $5770 \pm 100$ & $5710 \pm 180$ & \bm{$6150 \pm 180$}\\
 Test C & $750 \pm 820$  & $1090 \pm 660$&  $1160 \pm 540$ & $1040 \pm 760$ & \bm{$1370 \pm 800$}\\
 Test D & $5220 \pm 300$ & $5560 \pm 220$&  $5770 \pm 90$ & $5660 \pm 190$ & \bm{$6120 \pm 180$}\\
 Test E & $5270 \pm 290$  & $5570 \pm 210$ & $5770 \pm 100$ & $5660 \pm 220$ & \bm{$6140 \pm 150$}\\
 Test F & $780 \pm 860$  & $1160 \pm 570$   & $1120 \pm 580$ & $1140 \pm 870$ & \bm{$1390 \pm 750$}\\
 Test H & $130 \pm 290$  & $420 \pm 300$ & $210 \pm 220$ & $160 \pm 270$ & \bm{$700 \pm 560$}\\
 Test G & $5290 \pm 280$  & $5560 \pm 220$ & $5770 \pm 100$ & $5700 \pm 190$ & \bm{$6150 \pm 160$}\\
 \hline
\end{tabular}
\caption{Results on holdout tests for each of the tested approaches for Ant. Bolded values have the highest mean}
\label{table:mujoco_ant}
\end{table*}

\begin{figure}
     \centering
     \begin{subfigure}[b]{0.3\textwidth}
         \centering
         \includegraphics[width=\textwidth]{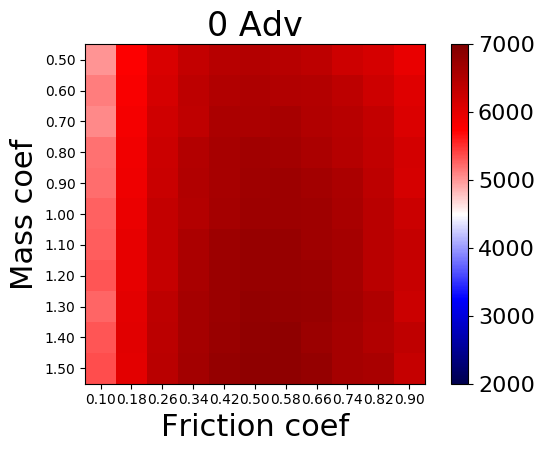}
     \end{subfigure}
     \begin{subfigure}[b]{0.3\textwidth}
         \centering
         \includegraphics[width=\textwidth]{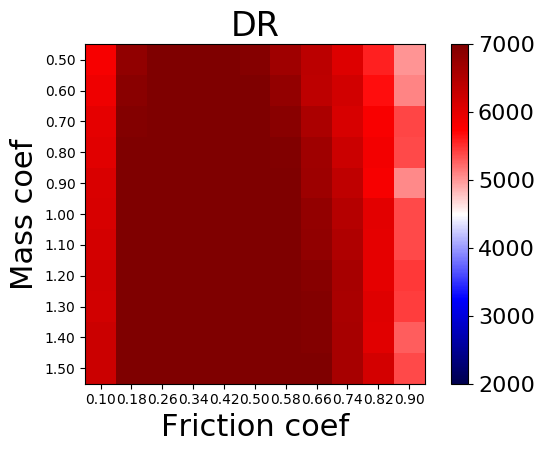}
     \end{subfigure}\\
     \begin{subfigure}[b]{0.3\textwidth}
         \centering
         \includegraphics[width=\textwidth]{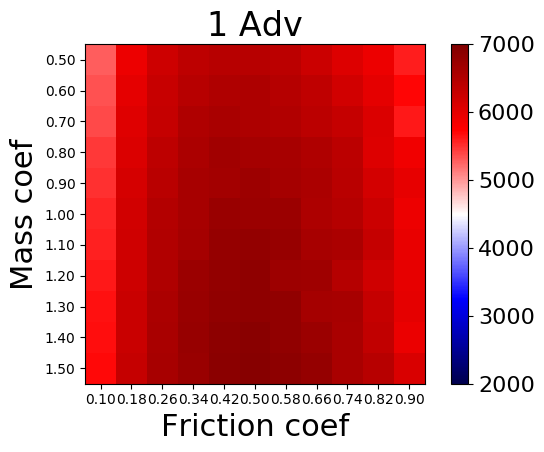}
     \end{subfigure} 
     \begin{subfigure}[b]{0.3\textwidth}
         \centering
         \includegraphics[width=\textwidth]{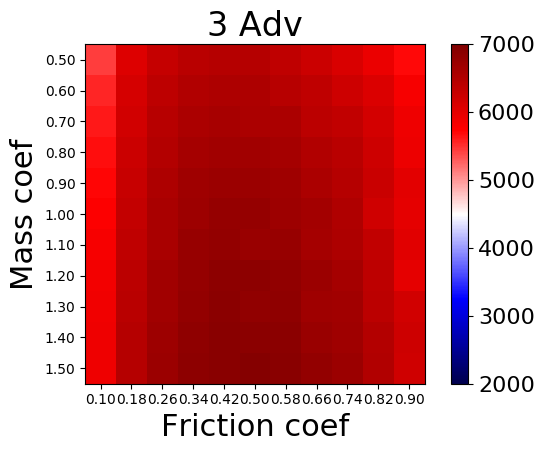}
     \end{subfigure}
     \begin{subfigure}[b]{0.3\textwidth}
         \centering
         \includegraphics[width=\textwidth]{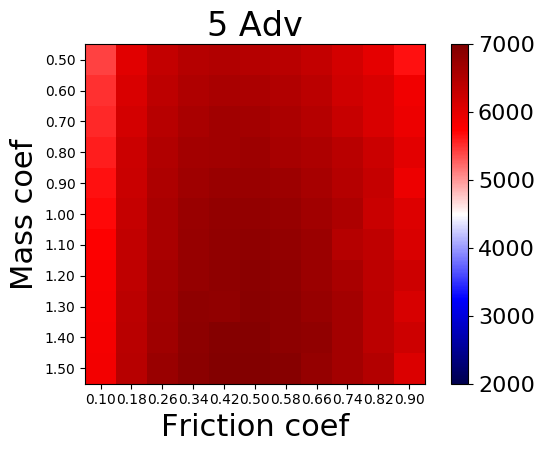}
     \end{subfigure}
        \caption{Ant Heatmap: Average reward across 10 seeds on each validation set (mass, friction) parametrization.}
        \label{fig:antheat}
\end{figure}

Finally, we place the heatmaps for Ant here for reference in Fig.~\ref{fig:antheat}. 

\section{Cost and Hyperparameters}
Here we reproduce the hyperparameters we used in each experiment and compute the expected run-time and cost of each experiment. Numbers indicated in $\{\}$ were each used for one run. Otherwise the parameter was kept fixed at the indicated value.

\subsection{Hyperparameters}
For Mujoco the hyperparameters are:
\begin{itemize}
    \item Learning rate: 
    \begin{itemize}
        \item $\{.0003, .0005\}$ for half cheetah
        \item $\{.0005, .00005\}$ for hopper
    \end{itemize}
    \item: Bounds on adversary action space: $\left[-0.25, 0.25\right]$ 
    \item Generalized Advantage Estimation $\lambda$
    \begin{itemize}
        \item $\{0.9, 0.95, 1.0\}$ for half cheetah
        \item $\{0.5, 0.9, 1.0\}$ for hopper and ant
    \end{itemize}
    \item Discount factor $\gamma = 0.995$
    \item Training batch size: $100000$
    \item SGD minibatch size: $640$
    \item Number of SGD steps per iteration: $10$
    \item Number of iterations: $700$
    \item We set the seed to 0 for all hyperparameter runs.
    \item The maximum horizon is 1000 steps.
\end{itemize}
For the validation across seeds we used 10 seeds ranging from 0 to 9. Values of hyperparameters selected for each adversary number can be found by consulting the code-base. All other hyperparameters are the default values in RLlib~\cite{liang2017rllib} 0.8.0.

\subsection{Cost}
For all of our experiments we used AWS EC2 c4.8xlarge instances which come with 36 virtual CPUs. For the Mujoco experiments, we use 2 nodes and 11 CPUs per hyper-parameter, leading to one full hyper-parameter sweep fitting onto the 72 CPUs. We run the following set of experiments and ablations, each of which takes 8 hours.
\begin{itemize}
    \item 0 adversaries
    \item 1 adversary 
    \item 3 adversaries
    \item 5 adversaries
    \item Domain randomization
\end{itemize}
for a total of 5 experiments for each of Hopper, Cheetah, Ant. For the best hyperparameters and each experiment listed above we run a seed search with 6 CPUs used per-seed, a process which takes about 12 hours. This leads to a total of $2 * 8 * 5 * 3 + 2 * 12 * 3 * 5 = 600$ node hours and $36 * 600 \approx 22000$ CPU hours. At a cost of $\approx 0.3$ dollars per node per hour for EC2 spot instances, this gives $\approx 180$ dollars to fully reproduce our results for this experiment. If the chosen hyperparameters are used and only the seeds are swept, this is $\approx 100$ dollars.

\end{document}